\documentclass{article}
\usepackage{ijcai23}

\usepackage{times}
\usepackage{soul}
\usepackage{url}
\usepackage[hidelinks]{hyperref}
\usepackage[utf8]{inputenc}
\usepackage[small]{caption}
\usepackage{graphicx}
\usepackage{amsmath}
\usepackage{amsthm}
\usepackage{booktabs}
\usepackage{algorithm}
\usepackage{algorithmic}
\usepackage[switch]{lineno}

\usepackage{multirow}
\usepackage{subfigure}
\usepackage{enumitem}
\usepackage{amsthm,amsmath,amssymb}
\usepackage{mathrsfs}

\title{KMF: Knowledge-Aware Multi-Faceted Representation Learning\\ for Zero-Shot Node Classification}

\author{
Likang Wu$^{1,2}$,
Junji Jiang$^3$,
Hongke Zhao$^4$,
Hao Wang$^{1,2,*}$
Defu Lian$^{1,2}$,
Mengdi Zhang$^5$\And
Enhong Chen$^{1,2}$
\affiliations
$^1$University of Science and Technology of China\and\\
$^2$State Key Laboratory of Cognitive Intelligence, China\and\\
$^3$Fudan University\and\\
$^4$Tianjin University\and\\
$^5$Meituan-Dianping Group
\emails
wulk@mail.ustc.edu.cn,
jjjiang22@m.fudan.edu.cn,
hongke@tju.edu.cn,
\{wanghao3,liandefu,cheneh\}@ustc.edu.cn,
zhangmengdi02@meituan.com
}
\begin{document}

\maketitle

\begin{abstract}
    Recently, Zero-Shot Node Classification (ZNC) has been an emerging and crucial task in graph data analysis. This task aims to predict nodes from unseen classes which are unobserved in the training process. Existing work mainly utilizes Graph Neural Networks (GNNs) to associate features' prototypes and labels' semantics thus enabling knowledge transfer from seen to unseen classes. However, the multi-faceted semantic orientation in the feature-semantic alignment has been neglected by previous work, i.e. the content of a node usually covers diverse topics that are relevant to the semantics of multiple labels. It's necessary to separate and judge the semantic factors that tremendously affect the cognitive ability to improve the generality of models. To this end, we propose a Knowledge-Aware Multi-Faceted framework (KMF) that enhances the richness of label semantics via the extracted KG (Knowledge Graph)-based topics. And then the content of each node is reconstructed to a topic-level representation that offers multi-faceted and fine-grained semantic relevancy to different labels. Due to the particularity of the graph's instance (i.e., node) representation, a novel geometric constraint is developed to alleviate the problem of prototype drift caused by node information aggregation. Finally, we conduct extensive experiments on several public graph datasets and design an application of zero-shot cross-domain recommendation. The quantitative results demonstrate both the effectiveness and generalization of KMF with the comparison of state-of-the-art baselines.
\end{abstract}

\section{Introduction}
The standard task of node classification~\cite{kipf2016semi,wu2023learning} predicts the labels of unlabeled ones on a graph, assuming that graph structure information reflects some affinities among nodes. In recent years, an emerging and more difficult task, zero-shot node classification~\cite{wang2021zero}, is proposed to predict the unlabeled nodes from unseen classes which are unobserved in the training process. This research paradigm is more conducive to catching up with the rapid growth of newly emerging classes in dynamic and open environments. For instance, the system of a scholar database tries to recognize a new research topic in citation networks without high cost and time-consuming annotations.

It is not easy to achieve the zero-shot learning for graph-structure data due to the complex relationships among special samples (nodes). Existing studies~\cite{wang2021zero,yue2022dual} mainly propose to use graph neural networks to construct node's representation, which associates prototypes and labels' semantics thus enabling knowledge transfer from seen to unseen classes. Different from the powerful GNNs used in traditional node classification, some generalization techniques are incorporated in this novel situation. Specifically, DGPN~\cite{wang2021zero} follows the principles of locality and compositionality for zero-shot model generalization. DBiGCN~\cite{yue2022dual} makes the semantic consistency from the perspective of nodes and classes.

\begin{figure}
    \centering
    \includegraphics[width=0.48\textwidth]{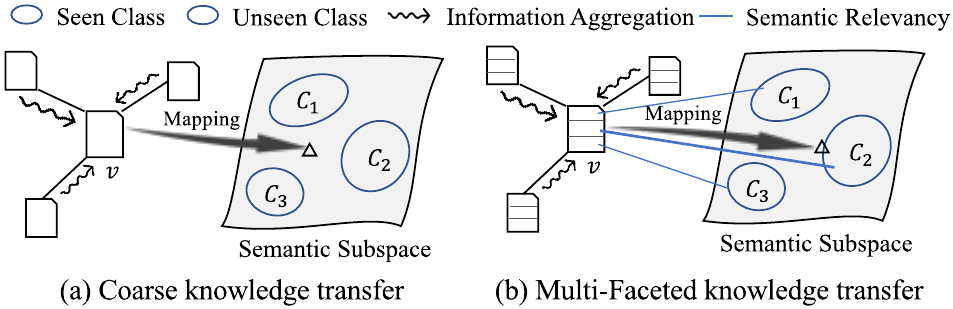}
    \caption{An overview of coarse knowledge transfer and multi-faceted knowledge transfer for zero-shot node classification. For semantic relevancy, a thicker dashed line indicates stronger relevancy.}
    \label{fig:intro}
\end{figure}

Although existing approaches classify nodes via establishing the mapping function from feature space to label semantic space, they ignore the multi-faceted and fine-grained semantic orientation in the knowledge transfer process. In fact, the content of a node covers diverse topics and is therefore relevant to the semantics of multiple labels, which is key to breaking the shackles of semantic confusion and promoting the generalization of models~\cite{wu2020multi}. As shown in Figure~1(a), when the model maps the representation vector of a node to the semantic space, it often encounters a dilemma that the distances between the vector and several label semantics are similar, leading to difficulty in classification. For the more discriminative strategy in Figure~1(b) proposed in this paper, we construct diverse sub-representations for the graph's nodes and build corresponding fine-grained semantic relevancy (orientation) with multiple labels' semantics. With these weighted semantic orientations, the model mines the major influencing factors to correctly judge the node's category, i.e., $C_2$ in Figure~1(b). 
Meanwhile, the explicit relevancy to the unseen class is also learned within training, which alleviates the bias that the trained model prefers seen classes. It can be seen that multi-faceted learning has the potential to acquire better classification accuracy and generalization in the ZNC task.

To achieve the high-quality zero-shot learning of multi-faceted knowledge transfer on graph data, there are some challenges that must be overcome on the path. First, it is hard to generate reasonable sub-representations which are aligned with diverse semantics of labels from the original content of nodes. We introduce a knowledge-aware topic mining algorithm  for this issue. Specifically, the labels (e.g., ``machine learning'', ``database'') are viewed as anchors on the given knowledge graph (we use ConceptNet~\cite{DBLP:conf/aaai/SpeerCH17} in our work). Since semantics are distributed on the knowledge graph in a structured and spatial form, we obtain several topic neighborhood sets based on the location of anchors. In this way, the original text of each node can be separated into multiple semantic sets associated with labels by matching the words of this node with topic sets. Then the topic-based node representation is embedded by a pre-trained language model, e.g., BERT~\cite{devlin2019bert,wu2023survey}, and a GNN for topological features mining. Hence we face the second challenge that the unified graph-based aggregation for the node usually leads to a node over-smooth problem, for the multi-faceted framework, we are also supposed to design a special differentiation enhancement mechanism. We regard the topic-based sub-representation as a deletable attribute, and propose a topic-view graph contrastive learning strategy to enhance the discrimination of nodes. Meanwhile, we find the node's aggregation would absorb information from the neighbors that are different from its own category. This diffusion makes prototype drift of each class in feature subspace, which further affects the alignment between semantic subspace and feature subspace in ZSL. To address this last challenge, we propose a geometric constraint loss that constrains the feature and semantic distribution collaboratively from both directions and distances, which can be easily integrated into our framework.

We conduct sufficient experiments on three real-world graph datasets. The quantitative results of zero-shot node classification demonstrate that our proposed Knowledge-Aware Multi-Faceted framework (KMF) gains state-of-the-art performance compared with a number of powerful baselines. Moreover, we design a zero-shot cross-domain recommendation application and the learned node representation shows a good performance on the items with new categories, which verifies the superiority of multi-faceted representation learning in improving the model's generalization capability.

The contributions of this paper are summarized as follows:
\begin{itemize}
\item To the best of our knowledge, we are the first to introduce the knowledge-aware mechanism into the ZNC problem, which depicts the topic-level multi-faceted semantic relevancy to promote cognitive ability.  
\item We adopt the topic-view graph contrastive learning and geometric constraint loss to address two representation issues caused by information aggregation of nodes, i.e., node over-smooth and prototype drift.
\item We experimentally demonstrate the effectiveness of KMF and further evaluate the generalization of produced representations on a downstream application.
\end{itemize}

\section{Related Work}
\subsection{Zero-Shot Learning}
Zero-shot learning (ZSL)~\cite{AliFarhadi2009DescribingOB,DBLP:conf/cvpr/LampertNH09,liu2020dual} tries to classify the instance belonging to unseen classes.
Existing ZSL methods are mainly limited to CV~\cite{WeiWang2019ASO} or NLP~\cite{WenpengYin2019BenchmarkingZT}. In the famous DAP~\cite{lampert2013attribute} model, an attribute classifier is first learned from source classes and then applied to unseen classes. The representative mapping-based methods~\cite{frome2013devise,norouzi2013zero} calculate the distances between object features and class features. Here class attributes or class semantic descriptions (CSDs)~\cite{lampert2013attribute} like ``wing'' and ``fur'' enable cross-class knowledge transfer. The classifier is trained to recognize these CSDs and then infer the CSDs of an unseen object to compare the results with those unseen classes to predict its label. Specifically, the classic and effective model ESZSL~\cite{romera2015embarrassingly} adopts a bilinear compatibility function to directly model the relationships among features, CSDs, and labels. WDVSc~\cite{wan2019transductive} adds different types of visual structural constraints to the prediction. Besides, some knowledge-based approaches are used to capture class relevancy recently~\cite{GCNZ,DGP,HVE,geng2021ontozsl}. Although ~\cite{wang2018rsdne} also considers the zero-shot setting in graph scenarios, they focus on graph embedding, not node classification. 
\subsection{Node Classification}
Early studies~\cite{zhu2003semi,zhou2003learning} generally handle node classification by capturing the graph structure information via shallow models.
Recently, GNN and its applications~\cite{scarselli2008graph,hamilton2017inductive,wu2021learning,wang2019mcne,han2023guesr}, especially graph convolutional networks~\cite{henaff2015deep}, have activated researchers' motivation because of the theory simplicity and model efficiency~\cite{bronstein2017geometric,chiang2019cluster}.
GNNs usually propagate information from each node to its neighborhoods with a designed filter at each aggregation layer.
Some powerful models attempt to advance this line including GAT~\cite{velivckovic2017graph}, GCNII~\cite{chen2020simple}, etc.
Nevertheless, the traditional line assumes that every class in the graph obtains some labeled nodes. The inability of recognizing unseen classes is one of the major challenges. Few studies explore this issue. DGPN~\cite{wang2021zero} creates locality and compositionality for zero-shot model generalization. DBiGCN~\cite{yue2022dual} jointly constrains representations of nodes and classes to make semantic consistency. However, they both neglect the necessity of multi-faceted features exacting for fine-grained knowledge transfer, which reduces the upper bound of the model's cognitive ability.
In this paper, we propose a multi-faceted framework for the zero-shot node classification problem from a knowledge-driven view.
\section{Methodology}
\subsection{Preliminary}
\textbf{Problem Formalization.} Let  $G = (V, E)$  denote a graph, where  $V$ denotes the set of  $n$  nodes  $\left\{v_{1}, \ldots, v_{n}\right\}$ , and  $E$  denotes the set of edges among these  $n$  nodes. Given the content of words group $W_v = \{ w_1, w_2, ..., w_{|v|} \}$ belonging to each node $v$. The class set in this graph is  $\mathcal{C}=\left\{\mathcal{C}^{s} \cup \mathcal{C}^{u}\right\}$  where  $\mathcal{C}^{s}$  is the seen class set and  $\mathcal{C}^{u}$  is the unseen class set satisfying  $\mathcal{C}^{s} \cap \mathcal{C}^{u}=\varnothing$. Supposing all training nodes are from seen classes $\mathcal{C}^{s}$, the goal of zero-shot node classification is to classify the rest of testing nodes whose label set is $C^{u}$.
\\
\textbf{Common-Sense Knowledge Graph.} We propose to use the common-sense knowledge graph to enhance the multi-faceted semantic representation. Some knowledge-enhanced document tasks~\cite{DBLP:conf/kdd/LiuLLWS021,IsmailHarrando2021ExplainableZT} assume that entities in textual data can be linked to corresponding entities in a KG. For example, news articles usually contain knowledge entities in Satori~\cite{yuqing2018Building} such as celebrities or organizations. These entities are usually the key message conveyed by the article. Actually, there are not so many conventional entities (proper or specific nouns, e.g., ``Christmas'', ``UCLA'', ``Paris'', ``Trump'') in graph data. We leverage ConceptNet~\cite{DBLP:conf/aaai/SpeerCH17}, a large-scale KG containing the statement of common-sense knowledge, to exact the common-sense entities (concepts) from the node content. The entities in KG represent concepts (words and phrases, e.g., $/c/en/time\_ser  ies$, $/c/en/pattern$, $/c/en/Data\_mining$) linked together by semantic relations such as $/r/IsA$, $/r/RelatedTo$, $/r/Synonym$. 

\subsection{Knowledge-Aware CSDs Construction}
For zero-shot node classification, a useful strategy is acquiring high-quality class semantic descriptions (CSDs) as auxiliary data, for transferring supervised knowledge from seen classes to unseen classes. Existing work~\cite{wang2021zero} proposes two types of CSDs: i) label-CSDs, the word/phrase embedding of class labels by pre-trained language models, i.e., the embedding of ``Physics", or ``Database". ii) text-CSDs, the document embedding generated from some class-related descriptions, like a paragraph describing this class. We argue that both of these two modes have their nonnegligible limitations. Label-CSDs are not informative enough and text-CSDs generally contain much noise. To overcome the weaknesses, we first propose KG-Aware CSDs, a high-quality semantic description with filtered external knowledge.

Considering the thin label- or text-CSDs are difficult to cover the rich semantics of representative topics, or domains~\cite{wang2021zero}. Our proposed solution promotes the coverage of each CSD for its corresponding domain. The underlying assumption is that words belonging to a certain topic are part of a vocabulary that is semantically related to its humanly-selected candidate label, e.g., a document about the topic of ``Database" will likely mention words that are semantically related to ``Database'' itself, such as ``data'', ``efficient'', and ``system''. To this end, we use ConceptNet to produce a list of candidate words related to the labels we are interested in, keeping only the English concepts for the English datasets. In other words, we generate a ``topic neighborhood" for each topic label which contains all the semantically related concepts/entities. Each label is presented by a single word or short phrase (e.g. ``Data Augmentation'', ``Information Retrieval''). We query ConceptNet via API\footnote{https://github.com/commonsense/conceptnet5/wiki} for entities that are directly connected to the given label. 

\begin{figure}
    \centering
    \includegraphics[width=0.5\textwidth]{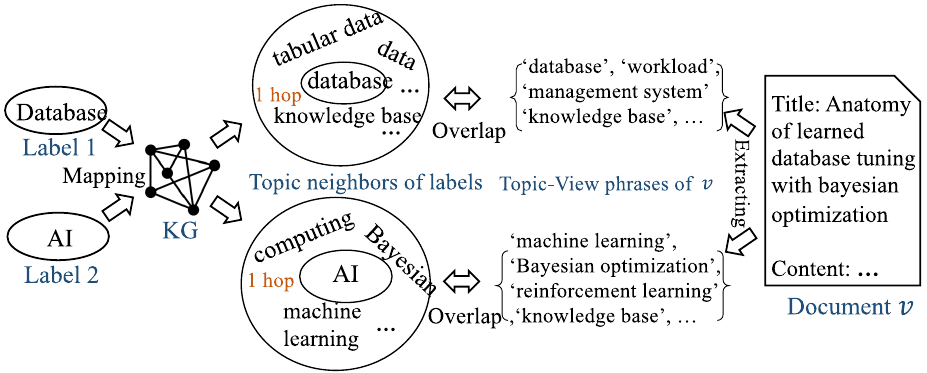}
    \caption{We generate the label's topic neighborhoods by querying a common-sense KG. For each node, its topic-view phrase sets with individual semantic orientations are extracted from the content.}
    \label{fig:CSD}
\end{figure}
The topic neighborhood is created by querying entities that are $N$ hops away from the label node as shown in Figure~\ref{fig:CSD}. Inspired by~\cite{IsmailHarrando2021ExplainableZT}, every entity is then given a score that is based on the Cosine similarity between the label and the entity computed using their embeddings with BERT. This score represents the relevance of any term in the neighborhood to the main label, and would also allow us to filter the noisy neighborhood. In the case of a label that has multiple tokens (e.g. the topic ``Arts, Culture, and Entertainment'') and cannot be linked into the KG, we just take the union of all word components' neighborhoods, weighted by the maximum similarity score if the same concept appears in the vicinity of multiple label components. We propose the coverage- and filtering-based strategy to vary the size of neighborhoods and reduce noise:
\begin{itemize}
\item \textbf{Coverage}. We vary the number of hops $R$. The higher $R$ is, the bigger the generated neighborhoods become;
\item \textbf{Filtering}. We filter out entities in each hop under Soft Cut (Top $P$ \%): we only keep the top $P$ \% entities in the neighborhood, ranked on their similarity score.
\end{itemize}

For each class $c \in \mathcal{C}$, the generated entities set is $\mathcal{T}_c=\{ w_1, w_2, ..., $ $w_{|\mathcal{T}_c|} \}$, where $|\mathcal{T}_c|$ denotes the size of topic neighborhood, $w_i$ is a word belonging to $\mathcal{T}_c$. The KG-Aware CSD of each class is the pooling of its topic neighborhood, 
{\setlength\abovedisplayskip{0.05cm}
\setlength\belowdisplayskip{0.05cm}
\begin{footnotesize}
\begin{equation}
S_c = \frac{1}{|\mathcal{T}_c|} \sum_{i=1}^{|\mathcal{T}_c|}a_i \cdot f_b(w_i) , w_i \in \mathcal{T}_c, c \in \mathcal{C},
\label{eq:CSD}
\end{equation}
\end{footnotesize}}
where $S_c \in \mathbb{R}^{d_b}$, $f_b(\cdot)$ denotes the function of BERT embedding. And, $a_i = 1.0 * (\alpha)^{r_i} $ is the attenuation weight of $w_i$, where $0 \leq \alpha \leq 1$ is a fixed attenuation coefficient, exponential $r_i \in \{0, 1, 2,..., R\}$ is the number of hops from $w_i$ to label $c$, i.e., the further a term is from the label vicinity, the lower is its contribution to the score.

\subsection{Multi-Faceted Node Representation Learning}
With the guidance of generated label topics, we design a framework of topic-level node representation learning to exact multi-faceted semantic orientations that transfer fine-grained knowledge from unseen classes to seen classes. 
\subsubsection{Topic-Level Node Representation}
Our high-precision model should capture multi-faceted semantic orientations behind the content. Meanwhile, the alignment with multiple semantics of labels is supposed to maintain within the training and testing procedure. We define the multi-faceted embedding of each node $v$ as a matrix $h_v \in \mathbb{R}^{|\mathcal{C}| \times d_b}$. Each row of $h_v$ is associated with a $d_b$ dimensional vector that illustrates the semantic relevancy to a certain label topic. Given the content $W_v = \{ w_1, w_2, ..., w_{|v|} \}$ of node $v$. To obtain the fine-grained embedding $h_v^{c}$, we first calculate the overlap between $W_v$ and $\mathcal{T}_c$. Following the assumption that the topic neighborhood covers the related concepts of this topic, the calculated overlap set $\mathcal{O}_v^c = \{ w_1, w_2, ... \}$ describes the focused point and orientation on topic $c$, for the content of node $v$. Similar to Eq. (\ref{eq:CSD}), the weighted embedding $h_v^c$ is acquired by a simple and efficient pooling as follows:
{\setlength\abovedisplayskip{0.05cm}
\setlength\belowdisplayskip{0.05cm}
\begin{footnotesize}
\begin{equation}
h_v^c = \frac{1}{|\mathcal{O}_v^c|} \sum_{i=1}^{|\mathcal{O}_v^c|}a_i \cdot f_b(w_i) , w_i \in \mathcal{O}_v^c, c \in \mathcal{C},
\label{eq:hvc}
\end{equation}
\end{footnotesize}}
where $h_v^c \in \mathbb{R}^{d_b}$. For the notation $a_i$ and $f_b(\cdot)$, please refer to the end of Section 3.2.

Considering the different distributions of semantic orientations, we regard the sizes of $\{ \mathcal{O}_v^c | c \in \mathcal{C} \}$ as weight coefficients while aggregating the compositionality of embeddings. So the compositional embedding of node $v$, $H_v = \sum_{c}^{\mathcal{C}} \frac{|\mathcal{O}_v^c|}{|\mathcal{O}_v|}  h_v^c$, where $\mathcal{O}_v = \bigcup_{c}^{\mathcal{C}} \mathcal{O}_v^c$. Expanding this equation, $H_v = \frac{1}{|\mathcal{O}_v|} \sum_{c}^{\mathcal{C}} \sum_{i=1}^{|\mathcal{O}_v^c|}a_i \cdot f_b(w_i)$. Since $\frac{1}{|\mathcal{O}_v|}$ is a global mean that has no impact on the weight distribution of semantic orientations, the actual impact factor $\sum_{i=1}^{|\mathcal{O}_v^c|}a_i$ of size-based coefficients approach is enough to incorporate the overlapping scale and relevance degree (the distance from label to overlap-words in the topic neighborhood) simultaneously.

Additionally, we further consider the too-large value deviation issue of $\sum_{i=1}^{|\mathcal{O}_v^c|}a_i$ for different semantic orientations. We replace  $\frac{|\mathcal{O}_v^c|}{|\mathcal{O}_v|} $ by a more smooth softmax coefficient $\frac{exp(|\mathcal{O}_v^c| / \tau)}{\sum_{c' \in \mathcal{C}} exp(|\mathcal{O}_v^{c'}| / \tau)}$, where the larger temperature parameter $\tau$ results in a more smooth probability distribution~\cite{DBLP:journals/corr/HintonVD15}. Therefore, the compositional embedding of node $v$ is defined as follows:
{\setlength\abovedisplayskip{0.05cm}
\setlength\belowdisplayskip{0.05cm}
\begin{footnotesize}
\begin{equation}
H_v = \sum_{c}^{\mathcal{C}} \frac{exp(|\mathcal{O}_v^c| / \tau)}{\sum_{c' \in \mathcal{C}} exp(|\mathcal{O}_v^{c'}| / \tau)}  h_v^c, v \in \mathcal{V}.
\label{eq:Hv}
\end{equation}
\end{footnotesize}}

To make the final representation capture the crucial topology information, we develop a concise and effective learning mode of gated message passing. Specifically, the compositional embedding $H_v$ is viewed as the initial state $H_{v}^{(0)}$ and the aggregation method for neighborhoods in each layer of message passing is presented as follows:
{\setlength\abovedisplayskip{0.08cm}
\setlength\belowdisplayskip{0.05cm}
\begin{scriptsize}
\begin{equation}
H_{v}^{(l+1)}=\sigma \Big(\xi^{(l)} W_H^{(l)} H_{v}^{(l)}+ (1-\xi^{(l)}) \frac{1}{|\mathcal{N}(v)|} \mathop{\sum}_{u \in \mathcal{N}(v)} W_H^{(l)} H_{u}^{(l)} \Big),
\label{eq:gcn}
\end{equation}
\end{scriptsize}}
where $\cdot ^{(l)}$ denotes related parameters in the $l$-th layer ($0 < l \leq L$), $\sigma = \frac{1}{1 + e^{-x}}$ is the sigmoid activation function, $0 < \xi^{(l)} \leq 1$ is a gated training parameter to control the updated degree, $W_{H}^{(l)} \in \mathbb{R}^{d_g^{(l)} \times d_g^{(l)}}$ is a weight matrix, i.e., the filter in graph neural networks. Along this line, the output node representation $H_{v}^G = H_{v}^{(L)} \in \mathbb{R}^{d_b}$ mines the tendentious compositionality of multi-faceted semantics and the structural knowledge of neighborhoods.

\subsubsection{Topic-View Graph Contrastive Learning}
The representations of nodes tend to be over-smoothing after several times of information aggregation, that is, the representations of interacting nodes converge to a similar vector, resulting in lower discrimination. In a zero-shot learning task, the less differentiated object representations are more likely to lead to the hubness issue in the semantic subspace. For our learning paradigm with multi-faceted semantics, we design a special topic-view graph contrastive learning module to enhance the node representation. We figure the node content is composed of several topics, and our basic motivation is to mask a few topics of nodes in the graph to produce a topic-view new graph that samples positive pairs and negative pairs. Compared with existing solutions for node over-smoothing, our topic-view contrastive learning can further enhance the importance of topic embeddings to the representation composition. We then specify the method details and contributions.

Similar to the salt-and-pepper noise in image processing, we add noise to nodes' topic 
components by randomly masking a fraction of components with zero vectors. For each node $v$ in the graph, we first sample a random vector $\tilde{m} \in \{ 0, 1 \}^{|\mathcal{C}|}$ where each dimension of it is drawn from a Bernoulli distribution independently, i.e., $\tilde{m}_c \sim Bern(1 - p_c^m), \forall c \in \mathcal{C}$. Then, the generated topic-view node features  $\tilde{H}_v$ is computed by:
{\setlength\abovedisplayskip{0.05cm}
\setlength\belowdisplayskip{0.05cm}
\begin{footnotesize}
\begin{equation}
\tilde{H}_v = \sum_{c}^{\mathcal{C}} \frac{exp(|\mathcal{\tilde{O}}_v^c| / \tau)}{\sum_{c' \in \mathcal{C}} exp(|\mathcal{\tilde{O}}_v^{c'}| / \tau)} \cdot h_v^c \circ \tilde{m}_c, v \in \mathcal{V},
\label{eq:mask}
\end{equation}
\end{footnotesize}}
where $\circ$ is the element-wise multiplication, $p_c^m$ is the probability of removing $h_v^c$ which reflects the importance of the $c$-th topic component to the node feature, $|\mathcal{\tilde{O}}_v^c| =  |\mathcal{O}_v^c| \cdot \tilde{m}_c$. We calculate the mask probability $p_c^m$ according to the normalization on the semantic orientation weights,
{\setlength\abovedisplayskip{0.05cm}
\setlength\belowdisplayskip{0.05cm}
\begin{small}
\begin{equation}
p_c^m=\min \left(\frac{s_{\max }^m-s_c^m}{s_{\max }^m-\mu_s^m} \cdot p_m, p_\tau\right),
\label{eq:pm}
\end{equation}
\end{small}}
where $s_c^m=\log \sum_{i=1}^{|\mathcal{O}_v^c|} a_i, s_{\max }^m$ and $\mu_s^m$ are the maximum and the average value of $s_c^m$, respectively, $p_m$ is the hyperparameter that controls the overall magnitude of feature augmentation, and $ p_{\tau}< 1$ is a cut-off probability, used to truncate the probabilities since extremely high removal probabilities will lead to overly corrupted node features.

After message passing, a symmetric entropy loss is used to optimize the distribution of each positive contrastive pair,
{\setlength\abovedisplayskip{0.05cm}
\setlength\belowdisplayskip{0.05cm}
\begin{scriptsize}
\begin{gather}
\mathcal{L}\left(H_{v}^G, \tilde{H}_{v}^G\right)=-\Big(\log \varphi\left(H_{v}^G, \tilde{H}_{v}^G\right)+\sum^{q \in \Phi_Q }_{q \neq v} \log \left(1-\varphi\left(H_{v}^G, H_{q}^G\right)\right)\Big), \nonumber \\
\mathcal{L}\left(\tilde{H}_{v}^G, H_{v}^G\right)=-\Big(\log \varphi\left(\tilde{H}_{v}^G, H_{v}^G\right)+\sum^{q \in \Phi_Q }_{q \neq v} \log \left(1-\varphi\left(\tilde{H}_{v}^G, \tilde{H}_{q}^G\right)\right) \Big),
\label{eq:nce}
\end{gather}
\end{scriptsize}}
where $Q$ negative embeddings $\Phi_{Q}$ obtained by random sampling from other nodes, $\varphi\left(H_{v}^G, \tilde{H}_{v}^G\right)=\sigma\left(H_{v}^G \cdot \tilde{H}_{v}^G\right)$ denotes the critic function and $\sigma$ is the sigmoid nonlinearity function. While training, the loss for contrastive learning is
{\setlength\abovedisplayskip{0.05cm}
\setlength\belowdisplayskip{0.05cm}
\begin{footnotesize}
\begin{equation}
\mathcal{L}_{cl} = \frac{1}{|V^s|} \sum_{v\in V^s} \left(\mathcal{L}\left(H_{v}^G, \tilde{H}_{v}^G\right) + \mathcal{L}\left(\tilde{H}_{v}^G, H_{v}^G\right) \right), 
\label{eq:losscl}
\end{equation}
\end{footnotesize}}
where $V^s$ is the set of seen nodes in training data. The training process makes node representations discriminative with both characteristics of alignment and uniformity, i.e., similar or homologous representations will be close, but the unique features of each one will be retained adequately.
What's more, for those nodes that miss some information about a topic, the contrastive loss for the positive pair encourages the attribute completion by exacting neighborhoods' features. For instance, some short texts may not explicitly mention all the topics involved, e.g., the data mining node with cross fields in the scholar network which is linked by papers' abstracts, corresponding to the node where we are masked out of some topics. By approximating the original node feature in the positive pair, the model forces the masked node to focus on completing the implied topics from the collaborative information of its neighborhoods. The node over-smoothing problem is also solved at the same time.

\subsection{Multi-Faceted Geometric Constraints}
Unlike conventional ZSL tasks in computer vision, there is a severe issue of prototype drift of the graph-based node representations, which is more obvious in our complex fine-grained learning. In general, the prototype of a class of samples can be represented by their centroid, i.e., the prototype $H_{c}^p$ indicates the mean value of representations $H_v^G$ belonging to class $c$. Since the generated node representation contains updated information from neighborhoods via message passing, the sample will inevitably incorporate feature information of other classes. The distribution of prototypes will drift, which disturbs the alignment between prototypes and label semantics. To alleviate this problem, we propose a multi-faceted geometric constraint. We correct the distribution of prototypes via different geometric properties from a spatial perspective. In the semantic subspace, we propose to achieve an alignment learning between prototypes and label semantics by two properties, that is, distance and relative direction,

{\setlength\abovedisplayskip{0cm}
\setlength\belowdisplayskip{0.05cm}
\begin{scriptsize}
\begin{gather}
\mathcal{L}_d = \mathop{argmin}\limits_{M_d^p} \frac{1}{|\mathcal{C}^s| (|\mathcal{C}^s|-1) /2} \Big( \sum_{0< i < j \leq |\mathcal{C}^s|} {\left| |M_{d,i,j}^p - M_{d,i,j}^l| - d_{\tau}\right| } \Big), \nonumber \\
\mathcal{L}_r = \mathop{argmin}\limits_{M_r^p} \frac{1}{|\mathcal{C}^s| (|\mathcal{C}^s|-1) /2} \Big( \sum_{0< i < j \leq |\mathcal{C}^s|} {\left| |M_{r,i,j}^p - M_{r,i,j}^l| - r_{\tau}\right| } \Big),
\label{eq:lossgc}
\end{gather}
\end{scriptsize}}
where $\mathcal{L}_d$ and $\mathcal{L}_r$ are constraint losses of distance and relative direction. $M_{d,i,j}^p = \sqrt{\sum_{k=0}^{d_b} (H_{i,k}^p - H_{j,k}^p)^2 }$ calculates the Euclidean distance between the prototype of class $i$ and $j$, $M_{d,i,j}^l = \sqrt{\sum_{k=0}^{d_b} (S_{i,k} - S_{j,k})^2) }$ calculates the Euclidean distance between the CSD of class $i$ and $j$, and the distance approximation is restricted by a hyperparameter threshold $d_{\tau}$. Similarly, $M_{r,i,j}^p = \frac{H_{i}^p \cdot H_{j}^p}{|H_{i}^p| \cdot |H_{j}^p|}$ calculates the Cosine similarity between the prototype of class $i$ and $j$, $M_{r,i,j}^l = \frac{S_{i} \cdot S_{j}}{|S_{i}| \cdot |S_{j}|}$ calculates the Cosine similarity between the CSD of class $i$ and $j$, and the relative direction approximation is restricted by a hyperparameter threshold $r_{\tau}$. During the training stage, the mapping network of node representations is encouraged to gradually align the consistency between prototypes and label semantics by the geometric constraints.

\subsection{Objective Function}
We regard the output layer of graph neural network in Eq. (\ref{eq:gcn}) as the map function which maps the node representation into our semantic space. In this space, we compute the prediction score $\zeta_{vc}$ of node $v$ w.r.t. a seen class $c \in \mathcal{C}^s$ as:
{\setlength\abovedisplayskip{0.05cm}
\setlength\belowdisplayskip{0.05cm}
\begin{small}
\begin{equation}
\zeta_{vc} = sim(H_v^G, S_c) = H_v^G \cdot S_c.
\label{eq:score}
\end{equation}
\end{small}}

The inference for unseen classes follows the operation as well as Eq. (\ref{eq:score}). The ground-truth label of node $v$ is $y_v \in \{0, 1\}^{\mathcal{C}^s}$, and the loss of node classification is defined as:
{\setlength\abovedisplayskip{0.05cm}
\setlength\belowdisplayskip{0.05cm}
\begin{small}
\begin{equation}
\mathcal{L}_c = - \frac{1}{|V^s|} \sum_{v\in V^s} \frac{1}{|\mathcal{C}^s|} \sum_{c \in \mathcal{C}^s} y_v(c)\mathrm{ln} (\hat{y}_v(c)), 
\label{eq:lossc}
\end{equation}
\end{small}}
where $\hat{y}_v(c) = \mathrm{softmax}(\zeta_{vc}) = \frac{exp(\zeta_{vc})}{\sum_{c' \in \mathcal{C}^s} exp(\zeta_{vc'})}$, $V^s$ is the set of seen nodes in training data. Finally, our overall model jointly optimizes the networks by integrating the classification loss (Eq. (\ref{eq:lossc})), the contrastive learning loss (Eq. (\ref{eq:losscl})), and the loss of geometric constraints (Eq. (\ref{eq:lossgc})) as follows:

\begin{equation}
\mathcal{L} = \mathcal{L}_c + \lambda_1 \mathcal{L}_{cl} + \lambda_2 (\mathcal{L}_d + \mathcal{L}_r),
\label{eq:loss}
\end{equation}

where $\lambda_1$ and $\lambda_2$ are trade-off hyper-parameters to control the influence of contrastive learning and geometric constraints. The construction of label topics and multi-faceted words assignment can be pre-computed in the offline stage.

\begin{table}[t]
	
	\centering
        \def\arraystretch{.8}
	\resizebox{.45\textwidth}{!}{
		\smallskip\begin{tabular}{ccccccc}
            \toprule
			\textbf{Dataset}  & \textbf{Nodes} & \textbf{Edges} & \textbf{Classes} & \textbf{Features} & \textbf{Word Embedding}\\
            \midrule
			DBLP & 17,725 & 52,914 & 4 & 1,617 & 1,024\\
			M10 & 10,310  & 77,218 & 10 & 932 & 1,024\\
			Cora$\_$E & 2,708 & 5,429 & 7 & 904 & 1,024\\
			\hline
		\end{tabular}
	}
 \caption{The statistics of datasets.}
	\label{table:Datasets}
\end{table}

\section{Experiments}
\subsection{Experimental Settings}
\textbf{Datasets.} We conduct experiments on three public real-world scholar datasets, DBLP, M10~\cite{pan2016tri}, and Cora$\_$E (Cora$\_$Enrich)~\cite{bojchevski2017deep}. The nodes, edges, and labels in these three represent articles, citations, and research areas, respectively. Compared with other accessible graph data, these datasets provide classification labels and raw text. We remove meaningless conjunctions and prepositions such as ``of'', ``for'', ``and'', etc. Following~\cite{pan2016tri}, we only keep words that occur more than 20 times in a dataset to avoid noise. Note that, due to the difference in information collection strategy and vocabulary, our datasets are not consistent with the datasets with the same names in previous studies, such as Cora. We also offer the word embedding matrix produced by BERT. The statistics of our processed datasets are listed in Table~\ref{table:Datasets}. 
\begin{table}[t]
    \centering
    \def\arraystretch{.9}
    
        \resizebox{0.9\columnwidth}{!}{
    \begin{tabular}{c|ccc|ccc}
        \toprule
        \multirow{2}{*}{Models} & \multicolumn{3}{c|}{Class Split I} & \multicolumn{3}{c}{Class Split II} \\
        \cmidrule(l){2-7}
          & Cora\_E & M10 & DBLP & Cora\_E & M10 & DBLP \\
        \midrule
                    RandG & 22.14 & 33.12 & 29.13 & 26.22 & 41.33 & 46.32     \\
                    DAP & 25.44 &	38.86 & 33.15 & 30.32	&47.12	&52.45        \\
                    ESZSL &  27.03	&37.14	&35.45$^*$ &  38.65	&55.42	&51.37   \\
                    WDVSc & 29.17	&38.15	&32.74 & 35.14	&45.14	&48.98 \\
                    DGPN & 33.65$^*$	&42.23$^*$	&35.17 & 46.35$^*$	&62.51$^*$	&56.12$^*$ \\
                    DBiGCN & 30.63	&38.79	&33.29 & 41.12	&57.84	&52.79 \\
                    KMF & \bf{36.15}	& \bf{44.13} &	\bf{38.65} &\bf{50.30}	&\bf{64.13}	&\bf{59.88} \\
        \midrule
                    Improve $\uparrow$ & +7.43\% & +4.50\% & +9.03\% & +8.52\% & +2.60\% & +6.70\% \\
   
        \bottomrule
        
    \end{tabular}
    
    }
    \caption{Zero-shot node classification accuracy (\%) on datasets, where $^*$ indicates the best result among baselines.}
    \label{tab:main-result}
    
\end{table}
\\
\textbf{Baselines.} We compare KMF with three types of baselines. i). RandomGuess (RandG), i.e., randomly guessing an unseen label, is introduced as the naive baseline. ii). The representative and powerful ZSL approaches for object recognition include DAP~\cite{lampert2013attribute}, ESZSL~\cite{romera2015embarrassingly}, and WDVSc~\cite{wan2019transductive}. 
We carefully adjust their implementation for adapting to our task. 
iii). The state-of-the-art models on zero-shot node classification, i.e., DGPN~\cite{wang2021zero}, DBiGCN~\cite{yue2022dual}. We do not use some KG models~\cite{GCNZ,HVE} since the number of categories is not enough to build a large network for relation learning, which leads to poor results in the existing work mentioned above.
\\
\textbf{Implementation Details.} 
In the experiments, \emph{ConceptNet} is used for capturing knowledge concepts of labels and topic neighborhoods. We tune the hyperparameters of KMF and baselines by hyperopt~\cite{bergstra2013hyperopt} to maintain fair competition. The embedding size of $e_i$ produced by BERT is 1,024. The radius of topic neighbors $R = \{0, 1, 2, 3\}$. We set the layer $L = [0, 1, ...,4]$ of GCN for compositional embedding, the dimension of hidden state and output $d_b = 1024$, the maximum of $\mathcal{N}(v)$ is set to 10, and the sampling number $Q = [0, 5, 10, 15, 20]$ of negative contrastive pairs in graph contrastive learning. The magnitude parameters $p_m, p_{\tau}$ $\sim$ uniform$[0.1, 0.5]$ in the topic mask sampling. In the geometric constraints, $d_r, r_{\tau}$ $\sim$ uniform$[0.1, 0.5]$ as well. For other hyperparameters, we set the attenuation coefficient $\alpha = 0.8$, temperature $\tau = 10$, filtering soft cut $\%P=25\%$, and trade-off $\lambda_{1,2}$ $\sim$ uniform$[0.1, 0.5]$ according to the feedback of experimental performance. We adopt Adam~\cite{Kingma2015AdamAM} with a learning rate of 0.001 to optimize our model. 

We design two seen/unseen class split settings like~\cite{wang2021zero}. We randomly adopt some classes as seen classes and the rest classes as unseen ones 3 times and calculate average results.
\textbf{Class Split I}: all the seen classes are used for training, and all the unseen classes are used for testing. The $[$ train$/$val$/$test $]$ class split for Cora$\_$E, M10, and DBLP are $[ 3 / 0 / 4 ]$, $[ 5 / 0 / 5 ]$, and $[ 2 / 0 / 2 ]$. 
\textbf{Class Split II}: the seen classes are partitioned into train and validation parts, and the unseen classes are still used for testing. The $[$ train$/$val$/$test $]$ class split for Cora$\_$E, M10, and DBLP are $[ 2 / 2 / 3 ]$, $[ 3 / 3 / 4 ]$, and $[ 2 / 2 / 2 ]$. Since the class number of DBLP is only 4, based on the setting of Class Split I, we randomly cut 20\% nodes belonging to unseen classes to create the validation set. 

\subsection{Zero-Shot Node Classification}
We conduct the zero-shot node classification task on our three datasets. The experimental results of Accuracy (ACC) are reported in Table~\ref{tab:main-result}. Overall, our proposed KMF model achieves the best performance among all baselines. KMF outperforms the most powerful baseline in each Class Split with 7.0\% and 5.94\% average relative accuracy improvements, compared with the increasing degrees of previous models, KMF expresses an obvious breakthrough in this difficult task. And KMF obtains the best ACC of 36.15\% and 50.30\% on the Cora$\_$E dataset under the settings of Class Split I and Class Split II respectively, which are great improvements on this dataset that the best scores among previous baselines are only 33.65\% and 46.35\%. Further analyzing the performance of baselines, we find that the performance of models using only graph neural networks or applying zero-shot learning framework is relatively close, and difficult to distinguish which learning mode is better. The fusion model of the above two modes, i.e., KMF, DBiGCN, and DGPN, performs well in most groups, reflecting this fusion paradigm's superiority. To sum up, the experimental results present both the effectiveness and strong generalization of learned knowledge-aware multi-faceted representation in zero-shot learning of nodes.

\begin{figure}
    \centering
    \includegraphics[width=.48\textwidth]{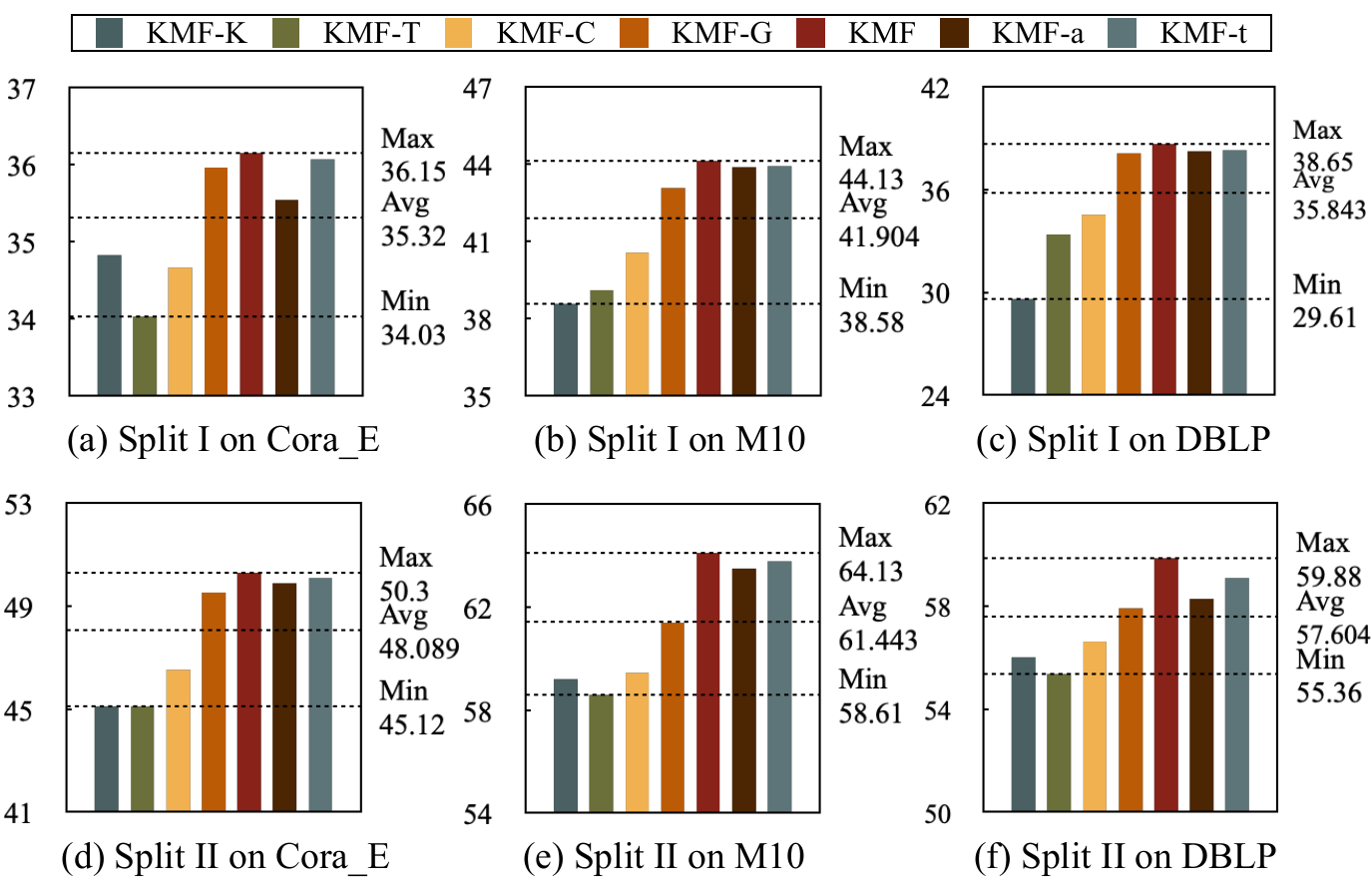}
    \caption{The ablation studies of KMF on three datasets under Class Split I and II. Each module contributes to the overall performance.}
    \label{fig:abl}
\end{figure}
\subsection{Ablation Studies}
To demonstrate the necessity of each proposed module, we carefully design and conduct the ablation experiments. Specifically, there are 6 sub-models KMF-K, KMF-T, KMF-C, KMF-G, KMF-a, KMF-t, and the complete model KMF. KMF-K refers to the variant that KMF uses the text-CSDs rather than the enhanced KG-Aware CSDs. KMF-T uses the KG-aware CSDs but removes the module of topic construction, which regards the embedding of node content as the node feature of a GCN. KMF-C, KMF-G does not consider the contrastive learning module, and the geometric constraint module, respectively. KMF-a and KMF-t are both more fine-grained variants. The former does not use the attenuation weight $a_i$ of $w_i$, while the latter removes the temperature parameter $\tau$. As shown in Figure~\ref{fig:abl}, all tested parts contribute to the final performance, which evidently demonstrates their effectiveness. What's more,  we find that the lack of multi-faceted representation modeling, i.e, KMF-T, leads to maximum performance loss. It shows that our motivation is the key to achieving accurate zero-shot node classification.

\subsection{Hyper-Parameter Tune-up}
\begin{figure}
    \centering
    \includegraphics[width=.48\textwidth]{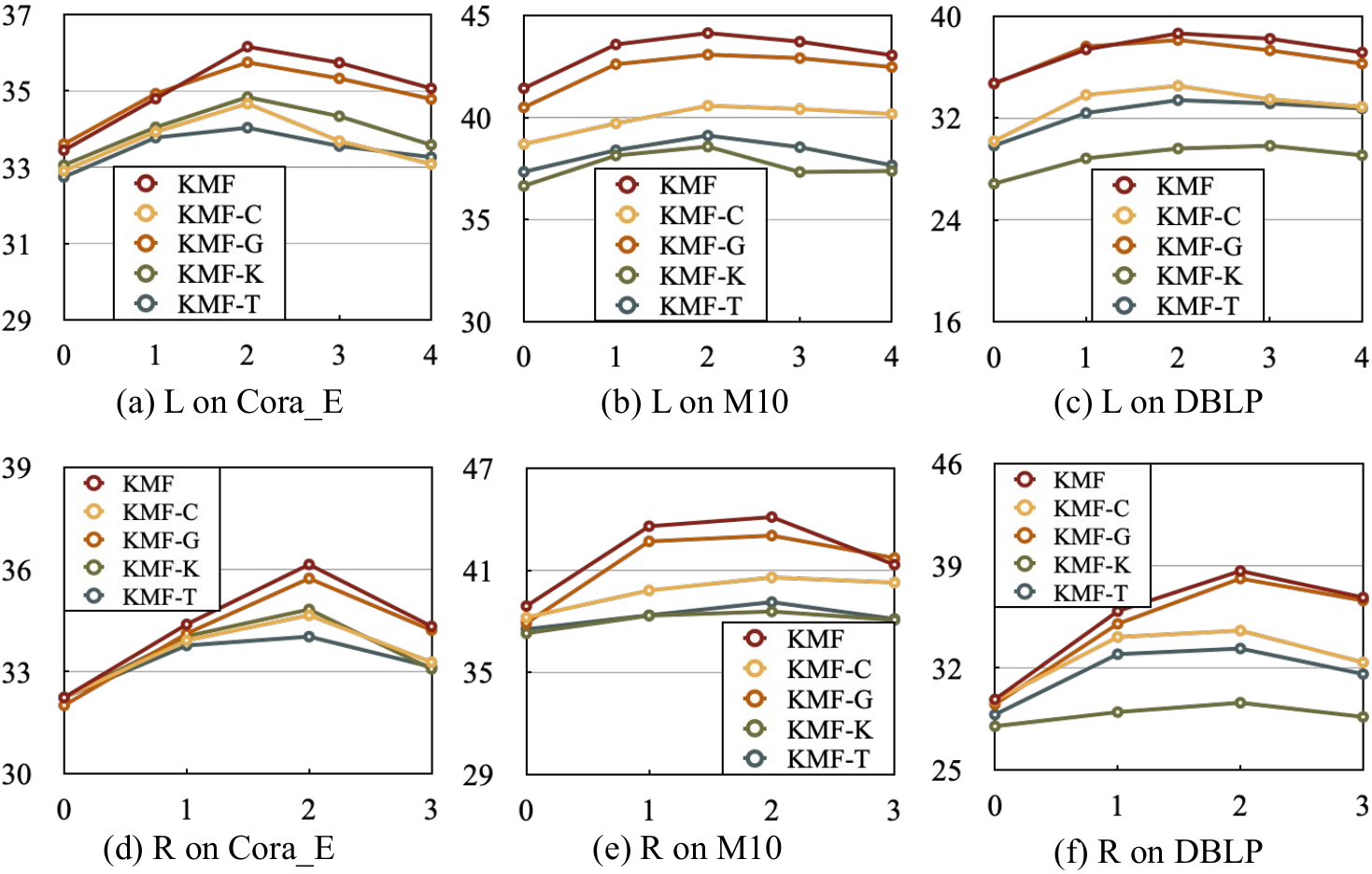}
    \caption{The tune-up experiments of the parameters, i.e., GCN layer $L$ and radius $R$ of topic neighborhoods under Class Split I.}
    \label{fig:hyper}
\end{figure}

In this section, we present the effect of two key hyper-parameters that are most important to our proposed model in the experiments: the radius $R$ of topic neighborhoods and the number $L$ of aggregation layers of GCN.
Here we also test the sub-models KMF-C, KMF-G, KMF-K, and KMF-T mentioned in Section 4.3. The best result of each hyper-parameter setting is illustrated in Figure~\ref{fig:hyper}. From the perspective of the global trend, with the increase of $R$ and $L$, the classification accuracies of our models in most settings improve steadily at firstly and gradually decrease after reaching the peak. To sum up, $L=2 $ can achieve good results in most cases, and $R=2$ is the same. The model with $R=2$ and $L=2$ is good and robust enough to deal with various situations. 

\begin{figure}[t]
  \centering
  \subfigure[KMF on $\mathcal{C}^{s}$]{
    \label{vis:a} 
    \includegraphics[width=0.31\columnwidth]{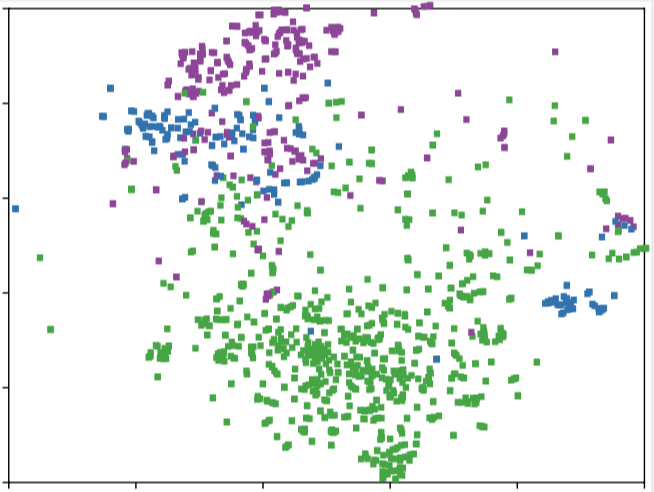}}
  \subfigure[KMF on $\mathcal{C}^{s} \cup \mathcal{C}^{u}$]{
    \label{vis:b} 
    \includegraphics[width=0.31\columnwidth]{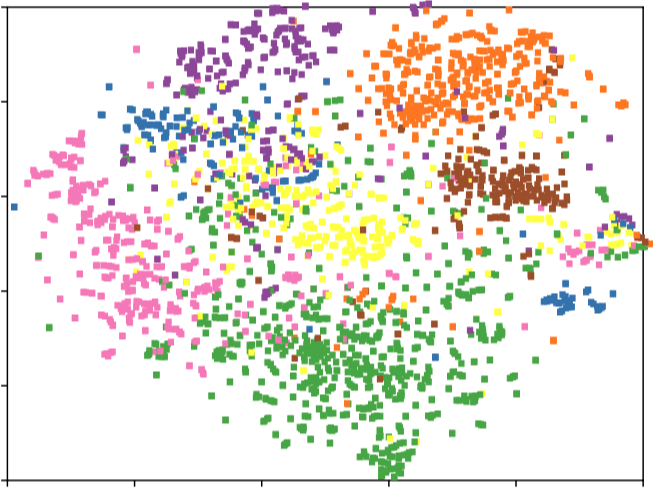}}
  \subfigure[DGPN on $\mathcal{C}^{s} \cup \mathcal{C}^{u}$]{
    \label{vis:c} 
    \includegraphics[width=0.31\columnwidth]{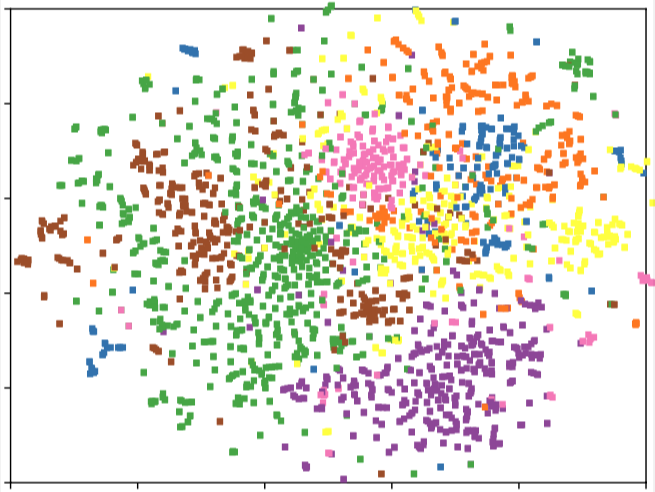}}
  \caption{The t$-$SNE visualization of representations of KMF and the representative approach DGPN on Cora$\_$E. }
  \label{fig:vis} 
\end{figure}
\subsection{Representations Visualization Analysis}
We try to demonstrate the intuitive changes of node representations after incorporating knowledge-aware multi-faceted representations. Therefore, like previous work~\cite{li2020learning,yu2022untargeted}, we utilize t-SNE~\cite{maaten2008visualizing} to transform feature representations (node embedding) of KMF and the representative approach DGPN into a 2-dimensional space to make an intuitive visualization. Here we visualize the node embedding of Cora$\_$E (actually, the change of representation visualization is similar in other datasets), where different colors denote different classes. According to Figure~\ref{vis:a} and \ref{vis:b}, the model obtains the cognitive ability to recognize unseen classes. And, there is a phenomenon that the visualization of KMF is more distinguishable than DGPN. The embedding learned by KMF presents a high intra-class similarity and divides nodes into different classes with distinct boundaries. On the contrary, the inter-margins of DGPN clusters are not distinguishable enough. 
\begin{table}[t]
    \centering
    \def\arraystretch{.85}
        \resizebox{0.9\columnwidth}{!}{
    \begin{tabular}{c|ccc|ccc}
        \toprule
        \multirow{2}{*}{Models} & \multicolumn{3}{c|}{Class Split I} & \multicolumn{3}{c}{Class Split II} \\
        \cmidrule(l){2-7}
          & AUC & HR@10 & MRR@10 & AUC & HR@10 & MRR@10\\
        \midrule
                    DAP & 62.15	& 39.1	& 22.3  &70.33	& 44.4 & 30.6      \\
                    ESZSL &  60.64 & 35.6 &	20.4 & 66.22 &	42.9 & 27.1   \\
                    WDVSc & 62.37 & 39.5 & 22.4 & 67.45	& 43.1 & 28.7  \\
                    DGPN & 63.12$^*$ &	40.9$^*$ & 23.1$^*$ & 71.31$^*$ &	46.1$^*$ & 31.5$^*$  \\
                    DBiGCN & 61.68	& 38.4 & 21.2 & 68.32 &	43.7 & 29.1  \\
                    KMF & \bf{65.19} &	\bf{41.8} & \bf{26.4} & \bf{73.27} & \bf{50.4} & \bf{33.3}\\
        \midrule
                    Improve $\uparrow$ & +3.28\% & +2.20\% & +14.29\% & +2.75\% & +9.33\% & +5.71\% \\
                    
        \bottomrule
    \end{tabular}
    }
    \caption{Zero-shot cross-domain recommendation results (\%) on DBLP, where $^*$ indicates the best result among baselines.}
    \label{tab:rec}
\end{table}
\subsection{Zero-Shot Cross-Domain Recommendation}
We design a zero-shot cross-domain recommendation application to further evaluate the generality of our representation. We choose DBLP due to its large number of nodes. Given a paper, the item-to-item task aims to recommend correct citations (papers) to it, which can be viewed as a link prediction. Each class denotes a domain in our work and the class $c$ is defined as a target (unseen) class. All edges (linked pairs) with the node belonging to $c$ are divided into the testing set, and others form the training set. For each training/testing linked pair, we generated 5/100 fake linked pairs, i.e., replacing its citation randomly. We report three measures of ranking quality: Area Under Curve (AUC), Hit Ratio (HR@10), and Mean Reciprocal Rank (MRR@10). Each class would be treated as the target class once, and we report the average result in Table~\ref{tab:rec}. Our KMF model achieves the best performance among all groups, which reflects the superior generality of our representation in such open-world downstream applications.

\section{Conclusion}
In this paper, we proposed a multi-faceted learning approach for zero-shot node classification, which first introduced the paradigm of knowledge-based feature enhancement into the emerging research area. Our model, KMF, expanded the semantic coverage of each label by constructing its knowledge-based topics. Then the overlap between the label topic and the node content was computed to generate a topic-level representation that offered multi-faceted semantic relevancy to multiple labels. We further considered the hard issue of prototype drift caused by message passing and designed a geometric constraint loss to align the semantic space of prototypes and labels. Sufficient comparison experiments on public datasets evaluated the effectiveness and generalization of KMF. In the future, we plan to introduce the generative model to further 
alleviate the model bias.

\section*{Acknowledgments}
The research was supported by grants from the National Natural Science Foundation of China (Grants No. 62202443, 72101176). The research was partially supported by Meituan.

\section*{Contribution Statement}
Likang Wu is responsible for the paper writing and experimental design. Junji Jiang conducts the experimental code and makes an equal contribution, and the other authors are responsible for the paper's ideas and review. Hao Wang is the Corresponding Author. Enhong Chen is the supervisor.

\bibliographystyle{named}
\bibliography{sample-base}

\end{document}